\documentclass[submission,copyright,creativecommons]{eptcs}
\providecommand{\event}{ICLP'21} 
\usepackage{breakurl}             
\usepackage{underscore}           
\usepackage{graphicx}

\title{Modeling and Solving Graph Synthesis Problems Using SAT-Encoded Reachability Constraints in Picat}
\author{Neng-Fa Zhou
\institute{CUNY Brooklyn College \& Graduate Center}
\email{zhou@sci.brooklyn.cuny.edu}
}
\def\titlerunning{Modeling and Solving Graph Synthesis Problems}
\def\authorrunning{N.F. Zhou}
\begin{document}
\maketitle
\begin{abstract}
  Many constraint satisfaction problems involve synthesizing subgraphs that satisfy certain reachability constraints. This paper presents programs in Picat for four problems selected from the recent LP/CP programming competitions. The programs demonstrate the modeling capabilities of the Picat language and the solving efficiency of the cutting-edge SAT solvers empowered with effective encodings.
  \end{abstract}

  \section{Introduction}
Picat \cite{PicatBook15} is a Prolog-like language that takes many features from other languages, including pattern-matching rules, functions, list/array comprehensions, loops, assignments, tabling for dynamic programming and planning, and constraint programming. These features make Picat a convenient modeling language for combinatorial problems, on a par with AMPL \cite{FourerG02}, OPL \cite{Hentenryck02}, and MiniZinc \cite{NethercoteSBBDT07}. As a logic language, Picat can often offer solutions that are as concise and elegant as the ones in ASP \cite{Brewka:2011:ASP}.

Picat supports constraint solving using different solvers, including CP (constraint programming), SAT (satisfiability), MIP (mixed integer programming), and SMT (SAT Modulo Theories). The last two decades have witnessed dramatic enhancement in SAT solvers' performance, thanks to inventions of techniques, from conflict-driven clause learning, backjumping, variable and value selection heuristics, to random restarts \cite{SATHAND,Bordeaux06,MalikZ09}. With findings of effective encodings \cite{Huang08,JeavonsP12,Knuth15,StojadinovicM14,TamuraTKB09,Walsh00,ZhouK17}, SAT has become a strong contendant for solving a wide range of constraint satisfaction and optimization problems (CSP).

Many CSPs involve synthesizing subgraphs that satisfy certain reachability constraints, including the constraint that ensures a cycle connecting all the vertices, as in the Hamiltonian cycle problem (HCP), and the constraint that ensures a strongly connected component.  For that reason, CP systems provide graph constraints for easing the modeling and solving of these problems \cite{gccat,minizinchandbook}. This paper addresses modeling graph synthesis problems in Picat and solving them using SAT. It describes programs for four problems selected from the recent LP/CP programming competitions, including the \textit{Roadrunner} problem from the 2019 competition, and the \textit{Masyu}, \textit{Shingoki}, and \textit{Tapa} problems from the 2020 competition. It also compares the programs with ASP programs for these problems. The Picat programs demonstrate the modeling capabilities of the Picat language and the solving efficiency of the cutting-edge SAT solvers empowered with effective encodings.

In \cite{DymchenkoM15}, programs in Picat are given for several Google Code Jam problems that utilize tabling and constraint programming. This paper can be considered as a sequel, which offers SAT-based solutions. The remainder of the paper is structured as follows: Section 2 briefly introduces constraint programming in Picat and describes the reachability constraints. Sections 3 through 6 give Picat programs for the four problems.\footnote{The complete programs are available at \url{https://github.com/nfzhou/lp-contest}} Section 7 describes the SAT encodings of the reachability constraints implemented in Picat, and presents the experimental results. Section 8 concludes the paper. The readers are assumed to be familiar with Picat. The overview chapter of the book \cite{PicatBook15} is a good quick start.

  \section{Picat and its Reachability Constraints}
  Picat provides four solver modules, \texttt{cp}, \texttt{mip}, \texttt{sat}, and \texttt{smt}, for modeling and solving CSPs. As a constraint programming language, Picat resembles CLP(FD) \cite{DincbasHSAGB88}: the operators \verb+::+ and \texttt{notin} are used for domain constraints, the operators \verb+#=+, \verb+#!=+, \verb+#>+, \verb+#>=+, \verb+#<+, \verb+#<=+, and \verb+#=<+ are used for arithmetic constraints, and the operators \verb+#/\+ (and), \verb+#\/+ (or), \verb+#^+ (xor), \verb+#~+ (not), \verb+#=>+ (if), and \verb+#<=>+ (iff) are used for Boolean constraints.\footnote{It is a tradition for CLP(FD) to prefix the sharp character to Prolog's operators to make constraint operators.}  Picat also supports table constraints and many global constraints.  

Recent additions into Picat include the \texttt{hcp} and \texttt{scc} constraints.  The \texttt{hcp($Vs$,$Es$)} constraint ensures that the \textit{directed} graph represented by $Vs$ and $Es$ forms a Hamiltonian cycle, where $Vs$ is a list of pairs of the form $\{V,B\}$, and $Es$ is a list of triplets of the form $\{V_1,V_2,B\}$. A pair $\{V,B\}$ in $Vs$, where $V$ is a ground term and $B$ is a Boolean (0/1) variable, denotes that $V$ is in the graph if and only if $B = 1$. A triplet $\{V_1,V_2,B\}$ denotes that $V_1$ is connected to $V_2$  by an edge in the graph if and only if $B = 1$.

  The \texttt{hcp} constraint has several variants. The \texttt{hcp($Vs$,$Es$,$K$)} constraint also forces the number of vertices in the graph to be $K$.  The \texttt{hcp\_grid($A$)} constraint ensures that the grid graph represented by $A$, which is a two-dimensional array of Boolean (0/1) variables, forms a Hamiltonian cycle. In a grid graph, each cell is directly connected orthogonally (i.e., horizontally and vertically, but not diagonally), to its neighbors. The \texttt{hcp\_grid($A$,$Es$)} constraint restricts the edges to $Es$, which consists of triplets. In a triplet $\{V_1,V_2,B\}$ in $Es$, $V_1$ and $V_2$ take the form $(R,C)$, where $R$ is a row number and $C$ is a column number, and $B$ is a Boolean variable. If $Es$ is a variable, then it is bound to the edges of the grid graph.  The \texttt{hcp\_grid($A$,$Es$,$K$)} constraint also enforces the number of vertices in the graph to be $K$.

  The \texttt{circuit} and \texttt{subcircuit} constraints are two classical graph constraints in CP. Let \texttt{L} be a list of domain variables \texttt{[X1,X2,...,Xn]}, where each variable corresponds to a vertex in the given graph and its domain represents the set of adjacent vertices.  A valuation of the domain variables satisfies the \texttt{circuit(L)} constraint if the subgraph represented by the valuation forms a Hamiltonian cycle. In the \texttt{subcircuit(L)} constraint, a vertex \texttt{Xi} is an in-vertex if \texttt{Xi} takes a value other than \texttt{i}. A valuation of the domain variables satisfies the \texttt{subcircuit(L)} constraint if the subgraph of all the in-vertices forms a Hamiltonian cycle.

  The following gives implementations of the \texttt{circuit} and \texttt{subcircuit} constraints using the \texttt{hcp} constraint.

  \noindent \rule{\textwidth}{1pt}
\begin{verbatim}
    circuit(L) =>
        N = len(L),
        L :: 1..N,
        Vs = [{I,1} : I in 1..N],
        Es = [{I,J,B} : I in 1..N, 
                        J in fd_dom(L[I]), 
                        J !== I, 
                        B #<=> L[I] #= J],
        hcp(Vs,Es).

    subcircuit(L) =>
        N = len(L),
        L :: 1..N,
        Vs = [{I,B} : I in 1..N, 
                      B #<=> L[I] #!= I],
        Es = [{I,J,B} : I in 1..N, 
                        J in fd_dom(L[I]), 
                        J !== I, 
                        B #<=> L[I] #= J],
        hcp(Vs,Es).
\end{verbatim}
\rule{\textwidth}{1pt}

\noindent The function \texttt{fd\_dom(V)} returns the domain of \texttt{V} as a list. In the implementation of \texttt{circuit}, as all the vertices are included in the subgraph, all the Boolean variables in the pairs of \texttt{Vs} are set to be 1. In the implementation of \texttt{subcircuit}, a vertex \texttt{I} is in the subgraph if and only if \verb+L[I] #!= I+, and there is an edge from vertex \texttt{I} to vertex {J} if and only if \verb+L[I] #= J+. Note that \texttt{subcircuit} cannot represent a subgraph that consists of a single vertex. This limitation is remedied by the \texttt{hcp} constraint.

The \texttt{scc($Vs$,$Es$)} constraint ensures that the \textit{undirected} graph represented by $Vs$ and $Es$ is strongly connected, where $Vs$ and $Es$ have the same forms as the arguments in the \texttt{hcp($Vs$,$Es$)} constraint. Note that the graph to be constructed is assumed to be undirected. If there exists a triplet $\{V_1,V_2,B\}$ in $Es$, then the triplet $\{V_2,V_1,B\}$ will be added to $Es$ if it is not specified. The \texttt{scc} constraint also has several variants. The \texttt{scc($Vs$,$Es$,$K$)} constraint also forces the number of vertices in the subgraph to be $K$. The \texttt{scc\_grid($A$)} constraint ensures that the grid graph represented by $A$ forms a strongly connected undirected graph, and the \texttt{scc\_grid($A$,$K$)} constraint also forces the number of vertices in the graph to be $K$.

\begin{figure}[h]
\begin{center}
\includegraphics[width=1.8in]{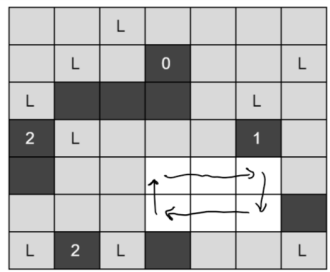}
\caption{\label{fig:roadrunner}A Road Runner instance and a solution}
\end{center}
\end{figure}

\section{Smarty Road Runner}
Given a grid map, where each cell is either an open area (called a white cell) or a hill (called a black cell), the goal of the Smarty Road Runner game is to install laser kits on some of the white cells such that the following rules are obeyed: 
\begin{itemize}
\item No two laser kits can shoot each other, meaning that no two laser kits can be installed in the same row or the same column, unless there is at least one black cell between them.
\item A black cell may have a number that indicates the number of laser kits installed in its quadrantal neighboring cells.
\item A white cell that is not covered by any laser beam is called a \textit{safe} cell. All safe cells must form one closed circuit for the Road Runner to run safely, meaning that the Road Runner can start running in any safe cell, follow to an adjacent safe cell, and arrive at the starting cell, traveling all safe cells without visiting any one twice.
\end{itemize}  
A laser kit, once installed, fires up horizontally and vertically, and laser beams do not stop until they reach a black cell or the edge of the grid. For example, Figure \ref{fig:roadrunner} shows an instance and a solution.\footnote{The image is taken from \url{https://github.com/lpcp-contest/lpcp-contest-2019}.}

The following gives a Picat program for the problem. 

\noindent \rule{\textwidth}{1pt}
\begin{verbatim}
import sat.

main([File]) =>
    read_data(File,MaxX,MaxY,Hill,Nums),
    Laser = new_array(MaxX,MaxY),
    Laser :: 0..1,
    Road = new_array(MaxX,MaxY),
    Road :: 0..1,
    foreach ($number(X,Y,Num) in Nums)
        sum([Laser[X1,Y1] : (X1,Y1) in [(X-1,Y), (X+1,Y), (X,Y-1), (X,Y+1)],
                            X1 >= 1, X1 =< MaxX,
                            Y1 >= 1, Y1 =< MaxY]) #= Num
    end,                        
    foreach (X in 1..MaxX, Y in 1..MaxY)
        (Hill[X,Y] == 1 ->
            Laser[X,Y] = 0,
            Road[X,Y] = 0
        ; 
            attacked_positions(Hill,X,Y,MaxX,MaxY,Ps),
            foreach ((X1,Y1) in Ps)
                Laser[X,Y] #=> #~Laser[X1,Y1],
                Laser[X,Y] #=> #~Road[X1,Y1]
            end,
            sum([Laser[X1,Y1] : (X1,Y1) in [(X,Y)|Ps]]) #= 0 #=> Road[X,Y]
        )
    end,
    K :: 1..MaxX*MaxY,
    hcp_grid(Road,_Es,K), 
    solve([$max(K)],Road),
    printf("safecircuitlen(%d).\n",K).
\end{verbatim}
\rule{\textwidth}{1pt}

\noindent The predicate \texttt{read\_data} reads the following items from an instance file:  \texttt{MaxX} and \texttt{MaxY} are, respectively, the number of columns and the number of rows, of the grid, counting from 1; \texttt{Hill} is a 2-dimensional 0/1 array, where an 1 entry indicates a black cell, and a 0 entry indicates a white cell; \texttt{Nums} is a list of terms of the form \texttt{number(X,Y,Num)}, which indicates that there must be \texttt{Num} lasers installed in \texttt{(X,Y)}'s neighbors.

The model uses a 2-dimensional array of Boolean variables, \texttt{Laser}, to indicate where lasers are installed, and another 2-dimensional array of Boolean variables, \texttt{Road}, to indicate the safe cells. The first \texttt{foreach} loop ensures that the required numbers of lasers are installed as specified in the input.\footnote{The \texttt{\$} symbol denotes that \texttt{number(X,Y,Num)} is a term, not a function call.}

The second \texttt{foreach} loop in the program enforces the relationship between the \texttt{Laser} and \texttt{Road} arrays. For each cell position \texttt{(X,Y)}, if the cell is a hill, then both \texttt{Laser[X,Y]} and \texttt{Road[X,Y]} are bound to 0, meaning that the cell can neither be a laser cell nor a road cell. Otherwise, the following actions are taken: (1) the call \texttt{attacked\_positions(Hill,X,Y,MaxX,MaxY,Ps)} finds a list, \texttt{Ps}, of the positions that are under attack by laser beams originating at \texttt{(X,Y)}; the \texttt{foreach} loop enforces that, if the cell at \texttt{(X,Y)} is a laser cell, then none of the positions in \texttt{Ps} can be a road or a laser cell; (3) the constraint
\begin{verbatim}
    sum([Laser[X1,Y1] : (X1,Y1) in [(X,Y)|Ps]]) #= 0 #=> Road[X,Y]
\end{verbatim}
ensures that, if no lasers are installed in any of the attacked positions, then  \texttt{(X,Y)} is a road cell.

The predicate \texttt{attacked\_positions} is defined as follows:
\begin{verbatim}
attacked_positions(Hill,X,Y,MaxX,MaxY,Ps) =>
    Ps = [(X1,Y) : X1 in X-1..-1..1, until(Hill[X1,Y] == 1)] ++
         [(X1,Y) : X1 in X+1..MaxX, until(Hill[X1,Y] == 1)] ++
         [(X,Y1) : Y1 in Y-1..-1..1, until(Hill[X,Y1] == 1)] ++
         [(X,Y1) : Y1 in Y+1..MaxY, until(Hill[X,Y1] == 1)].
\end{verbatim}
A list comprehension is utilized to collect the positions that are under attack by the laser beam in each of the four directions, and \texttt{Ps} is bound to the concatenation of the four lists created by the list comprehensions. Picat translates a list comprehension into a \texttt{foreach} loop that uses an assignment (\texttt{:=}) to accumulate values \cite{ZhouF17}. The \texttt{until($Condition$)} expression describes a terminating condition for the loop.

The \texttt{hcp\_grid} constraint ensures that all the road cells form a cycle.  It constrains \texttt{K} to be the total number of road cells:
\begin{verbatim}
    K #= sum([Road[X,Y] : X in 1..MaxX, Y in 1..MaxY])
\end{verbatim}  
The lower bound of \texttt{K} is 1, which ensures that there is at least one safe cell for the runner.

The objective of the problem is to find a subgraph such that \texttt{K} is maximized. The actual route is not required. If it were, it could be retrieved from the list of edges returned by \texttt{hcp\_grid}.

\section{Masyu}
\begin{figure}[h]
\begin{center}
\includegraphics[width=2.6in]{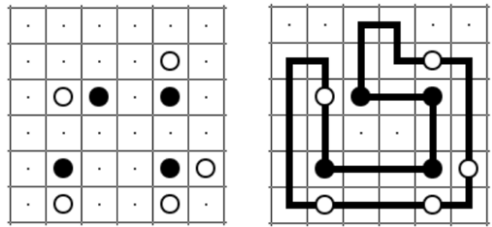}
\caption{\label{fig:masyu}A Masyu instance and its solution}
\end{center}
\end{figure}

\begin{figure}[h]
\begin{center}
\includegraphics[width=2.6in]{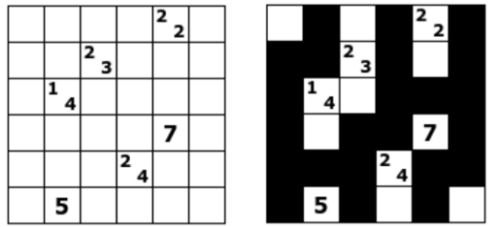}
\caption{\label{fig:tapa}A Tapa instance and its solution}
\end{center}
\end{figure}

Masyu is a logic puzzle played on a square grid. Some of the cells are marked with black circles, some are marked with white circles, and the rest are empty. The goal of the puzzle is to draw a single loop on the board, without crossings, that passes through all the black and white circles in the following fashion:
\begin{itemize}
\item White circles must be passed through in a straight line, but the loop must turn in the previous and/or the next cell.
\item Black circles must be turned upon, and the loop must travel straight through the next and the previous cell.
\item The loop can pass through any number of empty cells, as long as there are no crossings.
\end{itemize}
Figure \ref{fig:masyu} shows an instance and its solution.\footnote{The images shown in Figures \ref{fig:masyu}, \ref{fig:shingoki}, and \ref{fig:tapa} are taken from \url{https://github.com/alviano/lpcp-contest-2020}.}

The following gives a Picat program for the Masyu problem.

\noindent \rule{\textwidth}{1pt}
\begin{verbatim}
import util, sat.

main([File]) =>
    read_data(File,N,Board,Grid),
    hcp_grid(Grid,Es), 
    EMap = new_map(),
    foreach ({(R1,C1), (R2,C2), B} in Es)
        EMap.put({R1,C1,R2,C2}, B)
    end,
    foreach(R in 1..N, C in 1..N)
        if Board[R,C] == w then
            constrain_w(N,EMap,R,C)
        elseif Board[R,C] == b then
            constrain_b(N,EMap,R,C)
        end
    end,
    solve((Grid,values(EMap))),
    output(EMap).
\end{verbatim}
\rule{\textwidth}{1pt}

\noindent The predicate \texttt{read\_data} reads the following items from an instance file:  \texttt{N} is the grid size; \texttt{Board} is a 2-dimensional \texttt{N}$\times$\texttt{N} array that represents the configuration of the board, where white circles are denoted by the atom \texttt{w} and black circles are denoted by the atom \texttt{b}; \texttt{Grid} is a 2-dimensional \texttt{N}$\times$\texttt{N} array of Boolean variables, where the entries of white and black circles are all set to be 1, indicating that the loop passes through these cells, and the entries of empty cells are free variables.

The \texttt{hcp\_grid(Grid,Es)} constraint ensures that all the cells that are labeled 1 in \texttt{Grid} form a loop represented by a list of directed edges \texttt{Es}. A directed edge in \texttt{Es} has the form \texttt{\{(R1,C1),(R2,C2),B\}}, where \texttt{(R1,C1)} and \texttt{(R2,C2)} are two cell positions, and \texttt{B} is a Boolean variable that is equal to 1 if and only if the edge from \texttt{(R1,C1)} to \texttt{(R2,C2)} is included in the loop.

The loop must obey the rules when passing through black and white circles. In order to facilitate the use of \texttt{Es} returned by \texttt{hcp\_grid}, the implementation converts \texttt{Es} to a map. A triplet \texttt{\{(R1,C1),(R2,C2),B\}} is converted to a key-value pair, where the key is \texttt{\{R1,C1,R2,C2\}} and the value is \texttt{B}. The map makes it possible to retrieve the edge variable of a given edge in constant time.

The predicate \texttt{constrain\_w(N,EMap,R,C)}, which is defined below, constrains how the loop passes through the white cell at \texttt{(R,C)}. 
\begin{verbatim}
constrain_w(N,EMap,R,C) =>   
    Ps = [[(R,C-1), (R,C), (R,C+1), (R-1,C+1)],
          [(R,C-1), (R,C), (R,C+1), (R+1,C+1)],
          [(R-1,C-1), (R,C-1), (R,C), (R,C+1)],
          [(R+1,C-1), (R,C-1), (R,C), (R,C+1)],
          [(R-1,C-1), (R-1,C), (R,C), (R+1,C)],
          [(R-1,C+1), (R-1,C), (R,C), (R+1,C)],
          [(R-1,C), (R,C), (R+1,C), (R+1,C-1)],
          [(R-1,C), (R,C), (R+1,C), (R+1,C+1)]],
    constrain_paths(N,EMap,Ps).
\end{verbatim}
There must be a line in the loop that passes through the cell horizontally or vertically, and the loop must turn in the previous and/or the next cell. The predicate collects 8 path shapes into the variable \texttt{Ps}. Each path shape represents 2 paths, with the path shape itself representing one path and its reverse representing the other one. So, there are, in total, 16 possible ways that the loop passes through the white cell. For example, the path
\begin{tabbing}
  aaaa \= aaaa \= aaaa \= aaaa \= aaaa \= aaaa \= aaaa \= aaaa \= aaaa \= aaaa \= aaaa \= aaaa \= aaaa \= aaaa \kill
  \> \texttt{(R,C-1)} $\rightarrow$ \texttt{(R,C)} $\rightarrow$ \texttt{(R,C+1)} $\rightarrow$ \texttt{(R-1,C+1)}
\end{tabbing}
enters \texttt{(R,C)} from left, moves on to right, and turns up at \texttt{(R,C+1)}.

Let \texttt{B1} be the edge variable of \texttt{(R,C-1)} $\rightarrow$ \texttt{(R,C)}, \texttt{B2} be the edge variable of \texttt{(R,C)} $\rightarrow$ \texttt{(R,C+1)}, and \texttt{B3} be the edge variable of \texttt{(R,C+1)} $\rightarrow$ \texttt{(R-1,C+1)}. The above path is included in the loop if \verb+B1 #/\ B2 #/\ B3+ is satisfied.

Some of the path shapes in \texttt{Ps} may not be valid because they contain coordinates that fall outside of the grid. The predicate \texttt{constrain\_paths(N,EMap,Ps)} generates constraints to ensure that at least one of the valid path shapes in \texttt{Ps} occurs in the loop.

The predicate \texttt{constrain\_b(N,EMap,R,C)} constrains how the loop passes through the black cell at \texttt{(R,C)}. The loop must make a turn at \texttt{(R,C)}, and must travel straight through the next and the previous cell. In total, there are 8 possible ways that the loop passes through the black cell, among which the following is one:
\begin{tabbing}
  aaaa \= aaaa \= aaaa \= aaaa \= aaaa \= aaaa \= aaaa \= aaaa \= aaaa \= aaaa \= aaaa \= aaaa \= aaaa \= aaaa \kill
  \> \texttt{(R,C-2)}  $\rightarrow$ \texttt{(R,C-1)}  $\rightarrow$ \texttt{(R,C) $\rightarrow$ (R+1,C)}  $\rightarrow$  \texttt{(R+2,C)}
\end{tabbing}
The path enters \texttt{(R,C)} horizontally from left, and turns down at \texttt{(R,C)}.

The predicate \texttt{solve((Grid,values(EMap)))} retrieves the vertex variables from \texttt{Grid} and the edge variables from \texttt{EMap}, and labels these variables such that the constraints are satisfied.\footnote{The function \texttt{values(EMap)} returns a list of values in the key-value pairs of \texttt{EMap}.}

\section{Shingoki}
Shingoki is a logic puzzle, similar to Masyu, played on a square grid. Some of the cells are marked with black circles, some are marked with white circles, and the rest are empty. The goal of the puzzle is to draw a single loop, without crossings, that passes through all the black and white circles. White circles must be passed through in a straight line, and black circles must be turned upon. In Shingoki, each marked cell also has a number on it, which constrains the length of the two lines connected at the cell.

\begin{figure}[h]
\begin{center}
\includegraphics[width=2.6in]{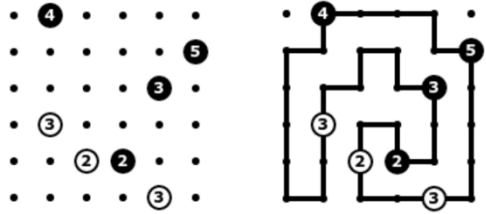}
\caption{\label{fig:shingoki}A Shingoki instance and its solution}
\end{center}
\end{figure}

Figure \ref{fig:shingoki} shows an instance of Shingoki and its solution. The cell marked with 4 is a black circle. The loop turns on the circle. The horizontal line connecting to it has length 3, and the vertical line connecting to it has length 1, making the total length equal to 4.

The following shows a Picat program for Shingoki.

\noindent \rule{\textwidth}{1pt}
\begin{verbatim}
main([File]) =>
    read_data(File,N,Board,Grid),
    hcp_grid(Grid,Es), 
    EMap = new_map(),
    foreach ({(R1,C1), (R2,C2), B} in Es)
        EMap.put({R1,C1,R2,C2}, B)
    end,
    foreach(R in 1..N, C in 1..N)
        if Board[R,C] = $w(Clue) then
            constrain_w(N,EMap,R,C,Clue)
        elseif Board[R,C] = $b(Clue) then
            constrain_b(N,EMap,R,C,Clue)
        end
    end,
    solve((Grid,values(EMap))),
    output(EMap).
\end{verbatim}
\rule{\textwidth}{1pt}

\noindent The \texttt{read\_data} predicate is the same as the one in the  Masyu program, except that \texttt{Board} is a 2-dimensional \texttt{N}$\times$\texttt{N} array that represents the configuration of the board, where an entry \texttt{w(Clue)} indicates a white circle with a clue number, and an entry \texttt{b(Clue)} indicates a black circle with a clue number. 

The predicate \texttt{constrain\_w(N,EMap,R,C,Clue)}, which is defined below, constrains how the loop passes through the white cell at \texttt{(R,C)}, which has the clue number \texttt{Clue}. 
\begin{verbatim}
constrain_w(N,EMap,R,C,Clue) =>
    Ps = [],
    foreach (D1 in 1..Clue-1, D2 = Clue-D1)
        V = [(R1,C) : R1 in R-D1..R+D2],
        P1 = [(R-D1,C-1)] ++ V ++ [(R+D2,C-1)],
        P2 = [(R-D1,C-1)] ++ V ++ [(R+D2,C+1)],
        P3 = [(R-D1,C+1)] ++ V ++ [(R+D2,C-1)],
        P4 = [(R-D1,C+1)] ++ V ++ [(R+D2,C+1)],
        H = [(R,C1) : C1 in C-D1..C+D2],
        P5 = [(R-1,C-D1)] ++ H ++ [(R-1,C+D2)],
        P6 = [(R-1,C-D1)] ++ H ++ [(R+1,C+D2)],
        P7 = [(R+1,C-D1)] ++ H ++ [(R-1,C+D2)],
        P8 = [(R+1,C-D1)] ++ H ++ [(R+1,C+D2)],
        Ps := [P1,P2,P3,P4,P5,P6,P7,P8|Ps]
    end,
    constrain_paths(N,EMap,Ps).
\end{verbatim}
There must be a line of length \texttt{Clue} passing through the cell, and the line can be split into two segments. The \texttt{foreach} loop considers all possible splits, with \texttt{D1} being the length of the first segment and \texttt{D2} being the length of the second segment (\texttt{D1+D2 = Clue}). The line can be either vertical or horizontal. The variable \texttt{V} refers to a list of cell positions in the vertical line, and the variable \texttt{H} refers to a list of cell positions in the horizontal line. Depending on how the loop turns at the ends of the line segments, there are, in total, 8 possible shapes of paths going through the cell. Each path shape represents two paths. So, there are, in total, 16 possible paths. The predicate collects all the path shapes into a variable \texttt{Ps} using assignments, and calls \texttt{constrain\_paths(N,EMap,Ps)} to ensure that at least one of the path shapes in \texttt{Ps} occurs in the loop.

The predicate \texttt{constrain\_b(N,EMap,R,C,Clue)} is defined similarly. As the loop turns on a black circle, two line segments connect orthogonally at the cell, and there are 4 possible connections, namely, $\lceil$, $\rceil$, $\lfloor$, and $\rfloor$. Counting in the possible ways the loop turns at the ends of the line segments, there are up to 16 possible shapes of paths going through a black cell.

\section{Tapa}
Tapa is another logic puzzle played on a square grid. Initially, some of the cells are filled with clue numbers and all others are empty. The goal of the puzzle is to color each of the cells black or white such that the following constraints are satisfied:
\begin{itemize}
\item The black cells form a single polyomino, connected orthogonally in a single group.
\item There are no 2$\times$2 black areas.
\item The clue numbers are respected.
\end{itemize}
A clue number indicates the size of a connected block of black cells in the surrounding neighbors, including diagonally connected neighbors. A cell can be filled with up to 4 clue numbers. If there are 2 or more clue numbers in a cell, then there must be at least 1 white cell between each 2 black blocks. The blocks can appear in any order. Figure \ref{fig:tapa} gives a Tapa instance and its solution. It can be checked that the solution respects all the clue numbers. For example, for the clue numbers 1 and 4 on the cell (3,2), row 3 and column 2, there is a block of size 1 at (4,3) and a block of size 4 occupying (2,1), (2,2), (3,1), and (4,1).

The following gives a Picat program for the Tapa problem.

\noindent \rule{\textwidth}{1pt}
\begin{verbatim}
main([File]) =>
    read_data(File,N,Board,Grid),
    scc_grid(Grid),
    foreach (R in 1..N-1, C in 1..N-1)
        Grid[R,C] + Grid[R,C+1] + Grid[R+1,C] + Grid[R+1,C+1] #< 4
    end,
    foreach (R in 1..N, C in 1..N, nonvar(Board[R,C]))
        neibs(N,R,C,NeibArr),
        constrain_blocks(Board[R,C],Grid,NeibArr)
    end,
    solve(Grid),
    output(Grid).

neibs(N,R,C,NeibArr) =>
    NeibArr = {(R1,C1) :  (R1,C1) in [(R-1,C-1), (R-1,C), (R-1,C+1), (R,C+1),
                                      (R+1,C+1), (R+1,C), (R+1,C-1), (R,C-1)],
                           R1 >= 1, R1 =< N, C1 >= 1, C1 =< N}.
\end{verbatim}
\rule{\textwidth}{1pt}

\noindent
The predicate \texttt{read\_data} reads the following items from an instance file:  \texttt{N} is the grid size; \texttt{Board} is a 2-dimensional \texttt{N}$\times$\texttt{N} array that represents the configuration of the board, where a non-variable entry indicates a list of clue numbers; \texttt{Grid} is a 2-dimensional \texttt{N}$\times$\texttt{N} array of Boolean variables, where a variable is labeled 1 if the cell is colored black, and 0 otherwise.

The \texttt{scc\_grid(Grid)} constraint ensures that all the black cells form a strongly connected component. The first \texttt{foreach} loop ensures there are no 2$\times$2 black areas. The second \texttt{foreach} loop ensures that the clue numbers are respected. For each entry \texttt{(R,C)}, if there are clue numbers filled in the cell, meaning that \texttt{nonvar(Board[R,C])} is true, then the program  ensures the existence of the given clue numbers of blocks in the neighbors.

The call \texttt{neibs(N,R,C,NeibArr)} binds \texttt{NeibArr} to an array of neighboring positions surrounding \texttt{(R,C)}. An array is more convenient than a list because the positions are circular.

The predicate \texttt{constrain\_blocks} is defined as follows:
\begin{verbatim}
constrain_blocks(Clues,Grid,NeibArr) =>
    findall_layouts(Clues,NeibArr,Layouts),
    Bs = [],
    foreach (Layout in Layouts)
        B :: 0..1,
        LayoutBs = [cond(Layout[I] == 1, Grid[R,C], #~Grid[R,C]) : 
                        I in 1..len(NeibArr), NeibArr[I] = (R,C)],
        B #= min(LayoutBs),
        Bs := [B|Bs]
    end,
    sum(Bs) #>= 1.
\end{verbatim}
The predicate \texttt{findall\_layouts(Clues,NeibArr,Layouts)} finds all the possible layouts of the blocks of the given sizes as indicated by \texttt{Clues}. A layout is an array where, an entry is 0 if the corresponding neighbor cell is free, and 1 if the corresponding neighbor cell is occupied by a block. For example, the cell at (3,2), row 3 and column 2, in Figure \ref{fig:tapa} has the following array of neighbors:
\begin{verbatim}
    NeibArr = {(2,1), (2,2), (2,3), (3,3), (4,3), (4,2), (4,1), (3,1)}
\end{verbatim}
The layout of the two blocks, of size 1 and size 4, respectively, given in Figure \ref{fig:tapa} is represented by the following layout:
\begin{verbatim}
    {1, 1, 0, 0, 1, 0, 1, 1}
\end{verbatim}
The conditional expression \verb+cond(Layout[I] == 1, Grid[R,C], #~Grid[R,C])+ is equal to \verb+Grid[R,C]+ if Layout[I] == 1, and \verb+#~Grid[R,C]+ otherwise. For each layout, a Boolean variable \texttt{B} is utilized to indicate if the blocks have the layout. The constraint \verb+sum(Bs) #>= 1+ ensures that the blocks have at least one of the layouts.

\section{Implementation and Experimental Results}
Several SAT encodings are available for the HCP problem \cite{HertelHU07,Prestwich03,VelevG09,Johnson14,Zhou20,Heule21}. If a graph has more than one vertex, then each vertex must have exactly one incoming edge and exactly one outgoing edge, and all the edges must form exactly one cycle. The degree constraints can be encoded easily. The focus of the encodings for HCP has been on how to ensure the reachability of all the in-vertices and ban sub-cycles. A common technique used is to impose a strict ordering on the vertices. The distance encoding chooses a vertex to serve as the starting vertex, and assigns a distinct distance to each of the in-vertices. If there is an edge from $v_i$ to $v_j$, then $v_j$'s distance is the successor of $v_i$'s, unless $v_j$ is the starting vertex. The successor function can be encoded in several different ways. In the implementation in Picat, the binary adder encoding is used, which has been found to be the most effective \cite{Zhou20}.

In contrast to HCP, no studies have been reported on encoding SCC into SAT. The challenge is centered on how to enforce the reachability of all the in-vertices. The satisfiability modulo acyclicity approach \cite{BomansonGJKS16} avoids encoding the constraint by performing reachability checking at solving time. The Picat implementation of \texttt{scc} employs an encoding, named \textit{tree encoding}, which utilizes the property that every strongly connected graph has at least one spanning tree. The tree encoding chooses a vertex as the root of a tree, chooses a parent for each non-root vertex, and assigns a distance to each vertex, with 0 being assigned to the root, and the distance assigned to each non-root vertex being 1 greater than that assigned to its parent.

For optimization problems, Picat uses branch-and-bound to optimize the objective. It first posts the problem as a constraint satisfaction problem, ignoring the objective. Once a solution is found, Picat uses binary search to find a solution with the optimum value. Each time the lower or upper bound is updated, Picat starts the SAT solver from scratch.

The rest of this section gives the execution times of the programs run using Picat version 3.1\footnote{\url{picat-lang.org}} on a Windows machine (Intel i7 3.30GHz CPU and 64G RAM). Picat provides C interfaces to several SAT solvers, with Maple\footnote{\url{http://sat-race-2019.ciirc.cvut.cz/}} as the default. The execution times reported below were obtained using Maple and Kissat version 1.0.3.\footnote{\url{http://fmv.jku.at/kissat/}}

For the sake of comparison, the execution times of the ASP programs for the problems provided by the LP/CP competition organizers run with Clingo version 5.5.0\footnote{\url{https://potassco.org/clingo/}} are also included. The reachability constraint is encoded neatly in ASP as follows:
\begin{verbatim}
    reach(V) :- start(V).
    reach(V) :- reach(U), edge(U,V).
    :- in(V), not reach(V).
\end{verbatim}
It ensures that there is a path from the starting vertex to every in-vertex. Clingo adopts the lazy approach to reachability checking \cite{BomansonGJKS16}.

Table \ref{tab:res} shows the execution times, which include both translation and solving times. The column Benchmark gives the instances, where the graph sizes are shown in parentheses. Picat(Kissat) is a clear winner.\footnote{Maple was the winner of the main track of the 2019 SAT Race, and Kissat was the overall winner of the 2020 SAT competition.} While Picat(Kissat) and Clingo demonstrate similar performance on Roadrunner and Tapa, Picat(Kissat) is more than 10 times as fast as Clingo on Masyu and Shingoki.

\begin{table}[tb]
  \begin{center}
    \caption{\label{tab:res}Execution times (seconds)}
    \begin{tabular}{|c|r|r|r|} \cline{1-4}
      Benchmark  & Picat(Kissat) & Picat(Maple) & Clingo \\ \cline{1-4}
Roadrunner6 (20$\times$20)   &  0.82  & 1.84 & 0.54 \\
 Roadrunner7 (20$\times$20)   &  5.05  & 42.52 & 1.33 \\
 Roadrunner8 (30$\times$30)   &  6.99  & 55.70 & 10.75 \\
 Masyu3 (30$\times$30)   &  1.62  & 6.94 & 22.37 \\
 Masyu4 (35$\times$35)   &  3.95  & 9.03 & 48.84 \\
 Masyu5 (40$\times$40)   &  6.21  & 11.72 & 96.78 \\
 Shingoki3 (31$\times$31)   &  2.89  & 7.11 & 27.28 \\
 Shingoki4 (36$\times$36)   &  5.26  & 8.59 & 57.42 \\
 Shingoki5 (41$\times$41)   &  8.62  & 14.23 & 111.62 \\
 Tapa3 (25$\times$25)   &  0.43  & 1.40 & 0.61 \\
 Tapa4 (30$\times$30)   &  0.62  & 3.74 & 0.97 \\
 Tapa5 (35$\times$35)   &  0.99  & 5.29 & 1.72 \\ \cline{1-4}
    \end{tabular}
  \end{center}
\end{table}

\section{Conclusion}
This paper presents the newly added graph constraints in Picat, and their use in modeling and solving four graph synthesis problems selected from the recent LP/CP competitions. The programs demonstrate the modeling capabilities of the Picat language. Picat provides several language constructs that are not present in standard Prolog, including functions, arrays, loops, list/array comprehensions, and assignments. With the graph constraints and these language constructs, Picat can serve as a powerful modeling language for various constraint solvers. The Picat programs compare favorably well in terms of conciseness with the ASP programs. As a general-purpose language, Picat has the advantage over many other modeling languages in handling I/O and integration with other software components.

The Picat programs also demonstrate the solving efficiency of the cutting-edge SAT solvers empowered with effective encodings. The bottleneck in solving the problems is ensuring the reachability of all the in-vertices. When the \texttt{hcp} and \texttt{scc} constraints are removed from the programs, all of the problem instances become trivial and can be solved in a flash. Picat follows the eager approach, first encoding constraints into SAT and then using a SAT solver to solve the encoding. This paper has shown that the eager approach is competitive with the lazy approach used in Clingo. With the advancement of SAT solvers and inventions of novel encoding methods, the eager approach will become ever more appealing.

The encodings for the \texttt{hcp} and \texttt{scc} constraints are the results of an extensive comparison study. Nevertheless, they are not meant to be optimal, and further research is warranted to find even better encodings. For optimization problems, such as the Road Runner problem, using an incremental SAT solver could lead to better performance. Furthermore, it is worthwhile to investigate translation of the graph constraints to SMT that supports checking of graph reachability at solving time.

\section*{Acknowledgement}
The author would like to thank the following people: H{\aa}kan Kjellerstrand for identifying many optimization opportunities in Picat's SAT compiler; Peter Bernschneider for sharing his Picat programs for the problems, which motivated the \texttt{hcp} and \texttt{scc} constraints and the programs presented in this paper; Mario Alviano for sharing his ASP encodings for the 2020 competition problems; Orkunt Sabuncu and Jose Morales for sharing the ASP encoding for the Road Runner problem. This work is supported in part by the NSF under the grant number CCF1618046.


\end{document}